\pdfoutput=1
\documentclass[11pt,preprint]{article}
\usepackage{adjustbox}
\usepackage{acl}

\usepackage{times}
\usepackage{latexsym}
\usepackage{multirow}
\usepackage{graphicx}
\usepackage{tipa}
\usepackage{tikz}
\usetikzlibrary{positioning, arrows.meta}
\usepackage{booktabs}
\usepackage{enumitem}
\usepackage{array}
\usepackage{textgreek}
\usepackage{siunitx}
\usepackage{stfloats}
\usepackage{booktabs,tabularx}
\usepackage{booktabs,makecell}
\usepackage{tabularx}
\usepackage{siunitx}
\usepackage{caption}  
\usepackage{subcaption} 
\usepackage{tipa}

\usepackage[T1]{fontenc}

\usepackage[utf8]{inputenc}

\usepackage{microtype}

\usepackage{inconsolata}

\usepackage{graphicx}

\title{Using the Mimi codec for metalinguistic representations}

\author{\rm
Artem Saloev$^1$,
Erin Pacquetet$^2$,
Nicolas Ballier$^1$\\
$^1$ALTAE, Universit{\'e} Paris Cit{\'e}, F-75013 Paris, France\\
$^2$SCIAM, 10 rue de Penthièvre, F-75008 Paris, France\\
    \texttt{artem.saloev@gmail.com, erin.pacquetet@sciam.fr,}\\ 
    \texttt{nicolas.ballier@u-pariscite.fr}
}

\begin{document}
\maketitle
\begin{abstract}
Codec-based audio language models are developing, but little explainability research has been dedicated to the representation of this type of speech tokenisation. In this paper, we focus on the dictionary of 2048 tokens used in Mimi's semantic token codebook, the neural codec of the Moshi language model \cite{defossez2024moshi}. We show that the ABX experiment carried out with Mimi fails to capture the mapping of the semantic tokens to phone realisations. By realigning Mimi's representations to the TIMIT corpus transcriptions \cite{garofolo1993timit}, we show that the 2048 tokens IDs of the semantic codebook map to quadphone, triphone, biphone, phone and subphone realisations. We used the TIMIT transcriptions as evidence of the validity of the allophone-based representations of these 80ms semantic token representation and examine some of the theoretical consequences for the tokenisation of speech at allophone and subphonemic level. 
\end{abstract}

\section{Introduction}
Codec-based audio language models have recently emerged as a way to apply Transformer-style sequence modelling to speech by first turning the waveform into a stream of discrete codec tokens. A neural audio codec compresses the signal using multiple codebooks, or layers (typically via residual vector quantisation, RVQ), yielding a small number of tokens per frame; a Transformer is then trained to predict token sequences, and a decoder reconstructs the waveform from those predictions.      
Working with discrete audio tokens brings benefits: stable long-context modelling, controllable bit-rate and latency (via tokens per second and the number of codebooks used), and, in some designs, faster generation by predicting coarse levels first and refining finer details in parallel. This type of technology is increasingly used for conversational agents or speech-to-speech translations \cite{labiausse2025hibiki} but little is known about how voices and speech features are encoded in these neural audio codecs. As conversational agents aim at real-time interactions and reducing latency, this has led to the design of codecs with limited frame rates (in our case study, 80 ms frames).
While this design is undoubtedly motivated by engineering considerations, the decision to discretise a naturally continuous acoustic signal raises a fundamental question : do the learned audio tokens partition the signal in ways that align with the categories posited by phonology (and, more broadly, phonetics)? Answering this question is central both to analysing these models’ internal representations and to assessing whether there is an alternative way to model the phon* level.
 
\begin{figure}
    \centering
    \includegraphics[width=1\linewidth]{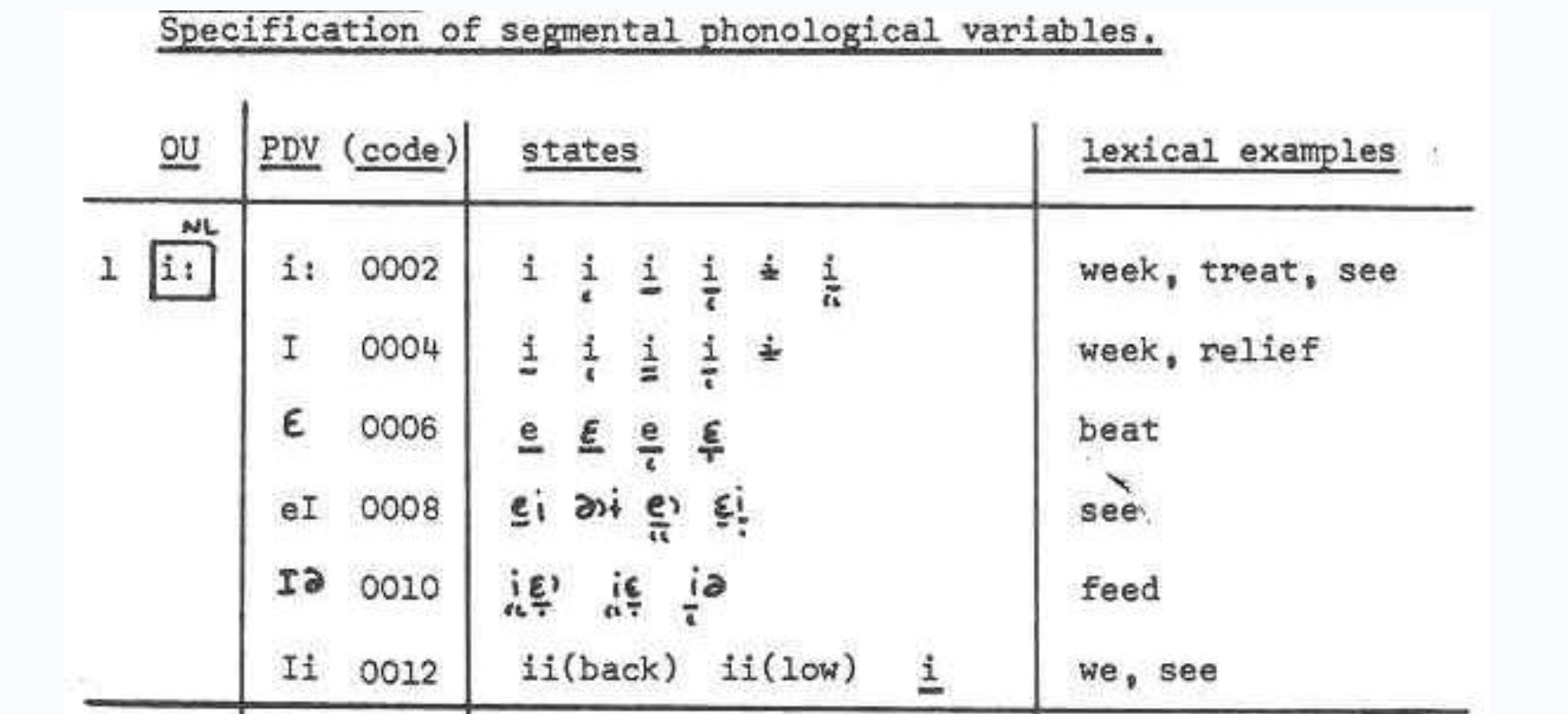}
    \caption{Levels of granularity in the Tyneside Linguistic Survey, from \cite{jones1985tyne}}
    \label{fig:DECTE}
\end{figure}

Our goal in this paper is to test how far the two representations coincide. We focus on Mimi, a recent streaming neural codec that produces multi-level token sequences suitable for spoken-dialogue modelling. Using TIMIT (the standard testbed for the phonetics of American English and its computational modeling), with its time-aligned phonetic transcriptions, we ask whether Mimi’s codebook tokens can be mapped onto the manually identified phonemes. This seemingly simple mapping raises immediate practical difficulties: token frames and phoneme boundaries are not naturally aligned; tokens have a fixed frame rate, whereas phoneme durations vary with context; and some information relevant to segment identity may be distributed across levels.
The audio language model Moshi generates and encodes speech as discrete audio tokens using a neural audio codec called ``Mimi''\footnote{\url{https://huggingface.co/kyutai/mimi}}. The Mimi codec quantizes audio waveforms into these tokens. It operates at a frame rate of 12.5 Hz (frames per second). This means that each ``temporal frame'' of audio processed by Moshi, and thus each time step for the Temporal Transformer, corresponds to an 80-millisecond (ms) slice of audio (1 second / 12.5 Hz = 0.08 seconds or 80 ms).
Moshi's design integrates the generation of both acoustic and semantic tokens within Mimi. Rather than having a completely separate semantic token codebook, Mimi's architecture is designed to distill semantic information into a specific part of its quantization process, effectively creating semantic-rich tokens.
Mimi discretizes waveforms into audio tokens from a self-supervised speech model, WavLM \cite{chen2022wavlm}, into tokens. This allows for streaming encoding and decoding of semantic-acoustic tokens. WavLM \cite{chen2022wavlm} was trained on English with Gigaspeech \cite{GigaSpeech2021} and on the 23 languages of the European Parliament with the VoxPopuli Dataset \cite{wang-etal-2021-voxpopuli}.
Our approach is bottom-up in spirit. We assume each 80 ms interval of audio is encoded by a single token in the Mimi codec encoder. We map codebook\_0 tokens (the semantic token codebook) to each 80 ms interval and re-align these intervals with the phone-aligned data of the TIMIT corpus. Because an 80 ms token may not capture the phone level, we cannot rely on the Phone Error Rate (PER) metric (an alternative reason being we have a dictionary of 2048 tokens for a 72 phone set). Based on standard studies of phone durations in English speech  \cite{keating1992phonetic, byrd1992preliminary, keating1994phonetic}, a plausible hypothesis is that an 80ms time frame is likely to capture triphones, biphones, phones and ``subphones'' (in particular, fragments of vowels). 
We partially replicate previous analyses of codecs that have been used for example with SpeechTokenizer \cite{zhang2023speechtokenizer}: we have used the TIMIT training data to match the Mimi semantic codebook token representations to TIMIT phonetic data transcriptions at phone level, word level and utterance level. We found the word-aligned transcriptions to be the most relevant to capture the gist of the semantic token representations, providing cogent representations of word allophone networks proposed in the early days of the TIMIT analyses. We have also used this available resource on English with the belief that the semantic codebook representations could be used to partially replicate previous phonetic analyses on the TIMIT data \cite{keating1992phonetic, byrd1992preliminary, keating1994phonetic,byrd1994relations} in the future, leading to extracting features for phone recognition tasks \cite{YOUNG1994369}, dialectology \cite{clopper2004some,clopper2004effects} and computational dialectology \cite{miller1996statistical}. For example, when investigating the phonetic realizations of Geordie (Newcastle English) in the 1960s for the Tyneside Linguistic Survey, linguists tried to annotate different levels of analysis. They considered as the major arch category the OU, the overall unit, and this more or less corresponds to the phonological level with the assumption that this abstraction corresponds to the class of realizations. In that respect, the mapping of the phonological category was not single-handedly over to the notion of phonetic realizations as diacritics, but was conceived as a form of several types of realizations. The PDV, the Putative Disystemic Variable, which assumes that a form of sub-categorization can be proposed to account for the different states, would correspond to individual realization, for example, as encountered in some specific lexical examples. So it might be the case that codec tokens could be very similar to the code that was associated in the transcription of the corpus that was coded on the basis of the four-digit code (see Figure \ref{fig:DECTE}). The (MIMI) codec tokens, if consistently encoding a portion of the acoustic signal bijectively could be used to encode the extreme level of granularity of the phonetic representation. Another potential application of this property would be the automatic annotation of code-switching. 

There are two main contributions in our paper. First, we establish a preliminary inventory of Mimi's ``semantic'' tokens observable in the TIMIT data with a method and code that can be replicated with the other 23 languages used to train the Mimi NAC. Second, we discuss the theoretical consequences of this type of speech tokenisation and evidence some of the potential benefits for the phonetic community of this type of ``semantic''  (indeed, allophonic) tokens.
The rest of the paper is organised as follows: Section \ref{prev_res} presents the main NACs that have been previously been investigated. Section \ref{methods} explains how we have analysed the Mimi codebook and how we organised the mapping of the semantic tokens with the TIMIT training data. Section \ref{results}
presents our results, Section \ref{discussion} discusses these results and Section \ref{future} suggests future research. Section \ref{conclusion} concludes.

\section{Previous Research}
\label{prev_res}

\subsection{Neural Audio Codec Investigation}
Previous research comparing NACs has tested neural codecs on text-to-sound task \cite{ye2025codec}, such as their ability to generate speech from texts using text-to-speech applying UTMOS as a metrics, a speech MOS (Mean Opinion Score) predictor \cite{saeki2022utmos}. They mostly investigate the interpretability of NAC at the decoding end, comparing the NAC outputs with human judgment. Other research focused on understanding how speech attributes like content, identity, and pitch are encoded within these codecs by proposing a two-step analysis and synthesis approach \cite{sadok2025bringing}. Though focusing on SpeechTokenizer and BigCodec, \cite{sadok2025bringing} partially discusses Mimi as one of four contemporary NACs studied for their ability to compress and decompress audio while maintaining a high level of quality. 
\cite{halimeh2025relation} investigates how the latent representations learned by neural speech codecs relate to quality but also demonstrates that there is not a direct, interpretable relationship between quantized latents and the qualities of the raw speech signal, further supporting the case that neural speech codec representations are less interpretable than raw signals. Lastly, \cite{mousavi2025discrete} proposed a typology of codecs and an across-the-board benchmark of codecs, extending the analysis of speech generation to audio and music generation.

\subsection{Mimi ABX experiment \cite{defossez2024moshi}}
When releasing the Mimi codec, \cite{defossez2024moshi} tested the potential phonetic representation of its semantic tokens using ABX discrimination tasks. The Phonetic Discriminability Evaluation, measured by the ABX error rate \cite{schatz13_interspeech}, was used to characterize the phonetic discriminability of the semantic token's representation space. This metric compares distances between embeddings of different instances of the same triphone (e.g., ``big'') versus a minimally different contrastive triphone (e.g., ``beg'') from the same speaker. A lower ABX error rate indicates better phonetic discriminability. 
We argue this procedure fails to capture what 80 ms of speech may represent phonetically. The ABX assumes phonemic representation but with an 80 ms slice of speech, we could be below the phoneme level in an elongated vowel, and with many consonants, we are at the cluster level. We believe the time frame associated to the Mimi tokens cannot be consistent with a phonemic representation, but is very likely to capture complex co-articulation phenomena: mutual influences of neighbouring sounds that are well-established in the phonetics/phonology literature. \footnote{Last, this moving windowed lens on the speech signal is very sensitive to phone alignment and the  Phonetic Discriminability Evaluation dataset used for the evaluation does not seem to have been manually checked.} The token interval may not correspond to the phone or word interval but still provide consistent representation  (encoding) of acoustic and phonetic properties of the aligned material (phone, diphone, triphone at the phonetic level; allophone, cluster or syllables at the phonological level).

\section{Material and Methods}
\label{methods}

\subsection{Models}
We used part of the inference code made public by Kyutai and extracted the 8 codebooks with the Mimi model available from hugging face.\footnote{\url{https://huggingface.co/kyutai/mimi.}} For this paper, we only investigated codebook\_0, the codebook supposedly corresponding to the ``semantic" tokens, specifically trained for this purpose \cite{defossez2024moshi}. As explained in \cite{defossez2024moshi}, during training, embeddings from WavLM \cite{chen2022wavlm} are distilled into a single vector quantizer which produces semantic tokens and that is combined with separate acoustic tokens for reconstruction. In other words, the model builders claim  that semantic information is more likely to be captured for this first codebook (codebook\_0): the tokens of this codebook are learnt by distillation from WavLM, a large-scale multi-lingual unlabelled audio dataset consisting of over 400k hours of audio in 23 languages collected from the European Parliament sessions, committee meetings, and other events. The audio tokens are running at 12.5Hz and with a bitrate of 1.1kbps. 
The semantic tokens are combined with separate acoustic tokens for reconstruction in the decoding phase. \cite{ballier26_speechprosody} confirmed that these codebooks, (codebook\_1) to (codebook\_7) partially captured prosodic features such as duration, pitch and glissando.

\subsection{Data}   
We used the official training and test partitions of the TIMIT Acoustic-Phonetic Continuous Speech Corpus \cite{garofolo1993timit}, a standard benchmark for phonetic and speech recognition research. The TIMIT dataset contains of speech recordings from 630 speakers covering eight major American English dialect regions, with each speaker reading 10 phonetically-balanced sentences selected to maximize coverage of phonemes and phonetic contexts.

\textbf{Data splits.} The corpus is divided into predefined training and test sets. The training set consists of 4,620 utterances from 462 speakers, while the test set includes 1,344 utterances from 168 speakers, ensuring balance across dialect regions and speaker demographics.

\textbf{Data structure.} Each utterance is provided as a 16kHz, 16-bit linear PCM waveform file, accompanied by time-aligned transcriptions at the sentence, word, and phoneme levels.
 
\subsection{Method} 
First, we encoded the audio files from the TIMIT training and testing data into Mimi codebooks\footnote{We used the HuggingFace implementation and the Kyutai github. These supplementary resources, as well as code for reproducibility (including realigned TextGrids and our codebooks), will be publicly released via Github upon publication.}. We then matched sequences of words and phones to their corresponding token encoding, focusing on the semantic tokens represented in codebook\_0. Because each 80\,ms interval of audio is encoded with a single token, the alignment between phones, words and token  rarely coincides. Some words are spoken throughout multiple 80 ms intervals, while some tokens are encoding for multiple phones at a time. Table \ref{table_alignment} shows an example of the time alignment mismatch between phone, word and token (codebook) level. 

\begin{table}[h!]
\begin{tabular}{@{}ccccc@{}}
\toprule
token & \begin{tabular}[c]{@{}c@{}}start \\ (in ms)\end{tabular} & \begin{tabular}[c]{@{}c@{}}end \\ (in ms)\end{tabular} & phone & word \\ \midrule
888 & 640 & 720 & epi|n|aw & now \\
495 & 720 & 800 & aw & now \\
1586 & 800 & 880 & aw|ix & \begin{tabular}[c]{@{}c@{}}now/\\ adjourned\end{tabular} \\ \bottomrule
\end{tabular}
\caption{Example of token to phone to word alignment}
\label{table_alignment}
\end{table}

Because the tokens produced are time-dependent, token time stamps do not align with phonetic transciptions and the pronunciation of words. We consequently decided to report three levels of analysis for the Mimi semantic tokens:  at phone level (interpreting the partial matching of token IDs with TIMIT phone set labels) at word level (interpreting the transcriptions of words in tokens) and at utterance level (trying to predict the different regions of the TIMIT data based on the transcriptions of utterances into tokens).

\subsection{Metrics} 
We analysed the relevance of the token sequences for transcriptions at phone level, word-level and utterance level. 
At the phone level, we could not use the more standard Phone Error Rate (PER) metric, which would compare a predicted phone sequence (from Mimi) to the gold standard TIMIT phonetic alignments. Instead, we replicated the method followed for the analysis of the SpeechTokenizer codec  \cite{zhang2023speechtokenizer} and computed the equivalent of the Phone-Normalized Mutual Information (PNMI) \cite{hsu2021hubert}. In other words, we investigate how pure the correspondence between TIMIT phone sequences and the time frames of the tokens. At the word-level, we used the training set of the TIMIT data and computed the percentage of words that were transcribed in the testing set with the same sequences of tokens as in the training data. At the utterance level, we used accuracy for the dialect classification task.

\section{Results}
\label{results}

\subsection{Phone level}
We first report the correspondences we observed between the assumed time frames of the semantic tokens and the most frequent co-occurring TIMIT phone or sequences of phones observed. 
In the training set of the TIMIT Corpus, 1793 tokens IDs were identified out of the 2048 dictionary IDs of the codebook. The utilization rate of the Mimi semantic codebook (87.54\%) is much more optimal than the one reported \cite{zhang2023speechtokenizer} for the EnCodec NAC \cite{defossez2023high}. Table~\ref{tab:TIMIdistrib} summarises the main correspondences between the semantic token time frames and the  TIMIT phones or phone sequences reported in the TIMIT corpus alignments.
\begin{table}[]
\centering
\begin{tabular}{ll}
\hline

\textbf{(majority) TIMIT correspondence} & \textbf{n}   \\
\hline
allophones/subphones            & 570 \\
diphones                        & 980 \\
triphones                       & 232 \\
quadphones                      & 11 \\
\hline
\end{tabular}
\caption{Distribution of TIMIT phone (sequences) matching Mimi semantic tokens}
\label{tab:TIMIdistrib}
\end{table}

Assuming the correspondence is between semantic token IDs and possible sequences of TIMIT phones, we computed the equivalent of the Phone-Normalized Mutual Information (PNMI) \cite{hsu2021hubert}, which computes the purity of the matching between the majority vote and the other candidates observed in the ``receptive field'' of the tokens. We obtained a value of 0.66, which ranks much above HuBERT (0.43), EnCodec (0.28) and slightly below SpeechTokenizer (0.71) but still similarly suggests that the semantic distillation process is effective \cite{zhang2023speechtokenizer}.

\subsection{Word-to-Token Re-transcription Predictability}
At word-level, using the word timestamps of the TIMIT corpus, we examined all the token retranscriptions of the words that were both present in the training set and in the testing set. For the 2,378 words present in the two sets, only 85 were transcribed differently in the testing set. Only 3.57 \% of the token sequences corresponding to the transcriptions of words were not predicted. It is likely that dialectal features could explain the observed variation in token sequences. Many frequent words having allophonic variations, such as words having weak forms (eg \textit{a, her, would}), and the very linguistic token chosen to design the comparable subcorpus SA, ``\textit{greasy}'' were found to have different token transcriptions in the training and testing sets.

\subsection{Utterance-Level Analysis}
 Assuming that the sequences of semantic tokens can be used as allophonic transcriptions of TIMIT data, we used the utterances read by all the speakers and tried to predict the dialect region of the speaker on the basis of the clusters of token IDs found in the transcription of each sound file. Assuming the eighth region is more difficult to predict as it corresponds to mobile speakers, we carried out a seven-class classification task.
We created a sparse matrix for the feature matrix that does not assume that all codebook\_0 vectors have the same length (Table \ref{tab:accents}). Each row in the dataset may have a codebook\_0 vector (token list) of a different length. The feature matrix is built by creating a sparse binary matrix where each unique token across the entire dataset becomes a column, and a value of ``1'' indicates that the token is present for a given sample. Sentences may have differing numbers of tokens, but the resulting matrix will always have the same number of columns (all unique tokens), with each row containing as many ``1''s as the number of tokens present for that sample. The method is more computer-intensive but robust to variable-length codebook\_0. We used a 80-20 split for a random forest model. The random forest model significantly outperformed the no-information rate, p $<$ .001, with an accuracy of 32.63\% (95\% CI [26.69\%, 39.01\%]) and a Cohen’s κ = 0.20.

\begin{table}
\centering
\begin{adjustbox}{width=0.48\textwidth}
\begin{tabular}[t]{lrrrrrrr}
\hline
  & \textbf{DR1} & \textbf{DR2} & \textbf{DR3} & \textbf{DR4 }& \textbf{DR5} & \textbf{DR6 }& \textbf{DR7}\\
\hline
DR1 & 2 & 0 & 0 & 0 & 0 & 0 & 0\\
DR2 & 8 & 17 & 12 & 4 & 3 & 6 & 10\\
DR3 & 1 & 7 & 12 & 5 & 4 & 4 & 11\\
DR4 & 0 & 4 & 6 & 13 & 8 & 0 & 4\\
DR5 & 4 & 0 & 1 & 16 & 16 & 0 & 2\\
DR6 & 2 & 1 & 1 & 0 & 3 & 5 & 1\\
DR7 & 2 & 11 & 8 & 2 & 5 & 3 & 12\\
\hline
\end{tabular}
\end{adjustbox}  \caption{Confusion matrix of the Random Forest prediction of the TIMIT dialect region  (Reference as columns, predictions as rows}
\label{tab:accents}
\end{table}

\subsection{Entropy-Based Token--Phoneme Analysis}
To refine the interpretation of these observations, we incorporated an entropy-based analysis of token--phoneme relationships using the TIMIT corpus. Importantly, TIMIT contains phonetically balanced sentences that are read by all speakers, including two shared utterances (SA1 and SA2). Although these sentences are not identical, they provide a controlled setting with consistent speaker coverage and comparable phonetic content.

The results reveal a strong asymmetry between the conditional distributions $P(\text{phoneme} \mid \text{token})$ and $P(\text{token} \mid \text{phoneme})$. Specifically, tokens exhibit low entropy and are strongly associated with a dominant phoneme, whereas each phoneme corresponds to a broad distribution of competing tokens.

\begin{figure}[h]
\centering
\includegraphics[width=\columnwidth]{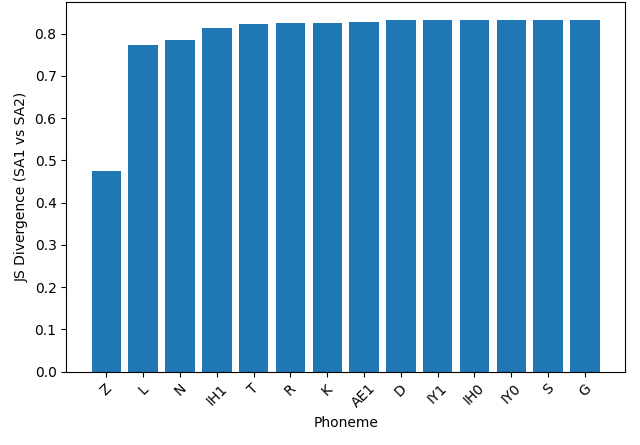}
\caption{Jensen--Shannon divergence between token distributions for shared sentences (SA1 vs SA2) across phonemes. High divergence values indicate substantial variation in token assignments across different phonetic contexts.}
\label{fig:js_divergence}
\end{figure}

To further investigate token stability, we compare token distributions for phonemes that occur in both SA1 and SA2. Despite being produced by the same set of speakers, we observe high Jensen--Shannon divergence between token distributions for many phonemes, often exceeding 0.8 (see figure \ref{fig:js_divergence}). This indicates substantial variation in token assignments across different phonetic contexts.

This many-to-one mapping provides quantitative support for the hypothesis that Mimi tokens encode finer-grained distinctions than phonemic categories. Rather than forming a discrete phoneme inventory, the first codebook partitions the acoustic space into clusters reflecting context-dependent realizations.

This variability cannot be fully explained by phonemic distinctions alone. Even when considering the same phoneme across comparable utterances, token distributions vary significantly, indicating sensitivity to contextual and coarticulatory factors. This result supports an allophonic interpretation of the tokens, where each phoneme is realized through multiple context-dependent variants encoded as distinct tokens.

\section{Discussion}
\label{discussion}

\subsection{Towards Subphonetic and Allophonic Representations}

The results suggest that Mimi's semantic codebook does not encode phonemes as discrete symbolic units, but rather captures finer-grained acoustic-phonetic patterns. This observation is consistent with the idea of ``bypassing transcription'' \cite{bird2020decolonising}, where linguistic structure is inferred directly from the speech signal without relying on predefined phonemic categories.

A useful parallel can be drawn with subphonetic modeling in traditional speech recognition. In Hidden Markov Model (HMM)-based systems \cite{rabiner2002tutorial}, context-dependent states (often referred to as senones) are used to capture variation in phoneme realizations across different contexts. These units do not correspond directly to phonemes, but rather to clustered acoustic realizations conditioned on phonetic environment \cite{hinton2012deep}, \cite{wang2023theory}.

Similarly, the Mimi tokens appear to function as context-dependent acoustic units. The entropy asymmetry and the high divergence observed across contexts indicate that a single phoneme corresponds to multiple token realizations, depending on coarticulation, prosody, and speaker-specific factors. It may suggest that the semantic codebook implicitly encodes a structured space of allophonic variation, and Mimi tokens can be interpreted as forming an emergent inventory of subphonetic units, situated between continuous acoustic representations and discrete phonological categories.

\section{Potential Applications for Phonetic Modeling}
\label{applications}
In this section, we compare the speech tokenization operated by Mimi to previous subphonetic modeling representations, especially for context-dependent subphonetic representations \cite{hwang1992subphonetic}.

\subsection{Revisiting Allophone Networks}
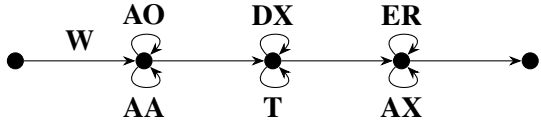
\begin{figure}[h]
\centering
\begin{tikzpicture}[
  node distance=1.7cm,
  every node/.style={font=\bfseries},
  state/.style={circle, draw, minimum size=6pt, inner sep=0pt, fill=black},
  phoneme/.style={font=\sffamily\bfseries},
  >={Stealth}
]

\node[state] (start) {};
\node[state, right of=start] (s1) {};
\node[state, right of=s1] (s2) {};
\node[state, right of=s2] (s3) {};
\node[state, right of=s3] (end) {};

\draw[->] (start) -- node[above]{W} (s1);
\draw[->] (s1) -- (s2);
\draw[->] (s2) -- (s3);
\draw[->] (s3) -- (end);

\draw[->, looseness=8, out=135, in=45] (s1) to node[above]{AO} (s1);
\draw[->, looseness=8, out=225, in=315] (s1) to node[below]{AA} (s1);

\draw[->, looseness=8, out=135, in=45] (s2) to node[above]{DX} (s2);
\draw[->, looseness=8, out=225, in=315] (s2) to node[below]{T} (s2);

\draw[->, looseness=8, out=135, in=45] (s3) to node[above]{ER} (s3);
\draw[->, looseness=8, out=225, in=315] (s3) to node[below]{AX} (s3);

\end{tikzpicture}
\caption{Allophone network for the word \textit{water} \cite{cohen1987studies}}
\label{AllophoneNetwork}
\end{figure}

\begin{figure*}[htbp]
    \centering
    \begin{subfigure}[t]{0.48\textwidth}
        \centering
        \includegraphics[width=\linewidth]{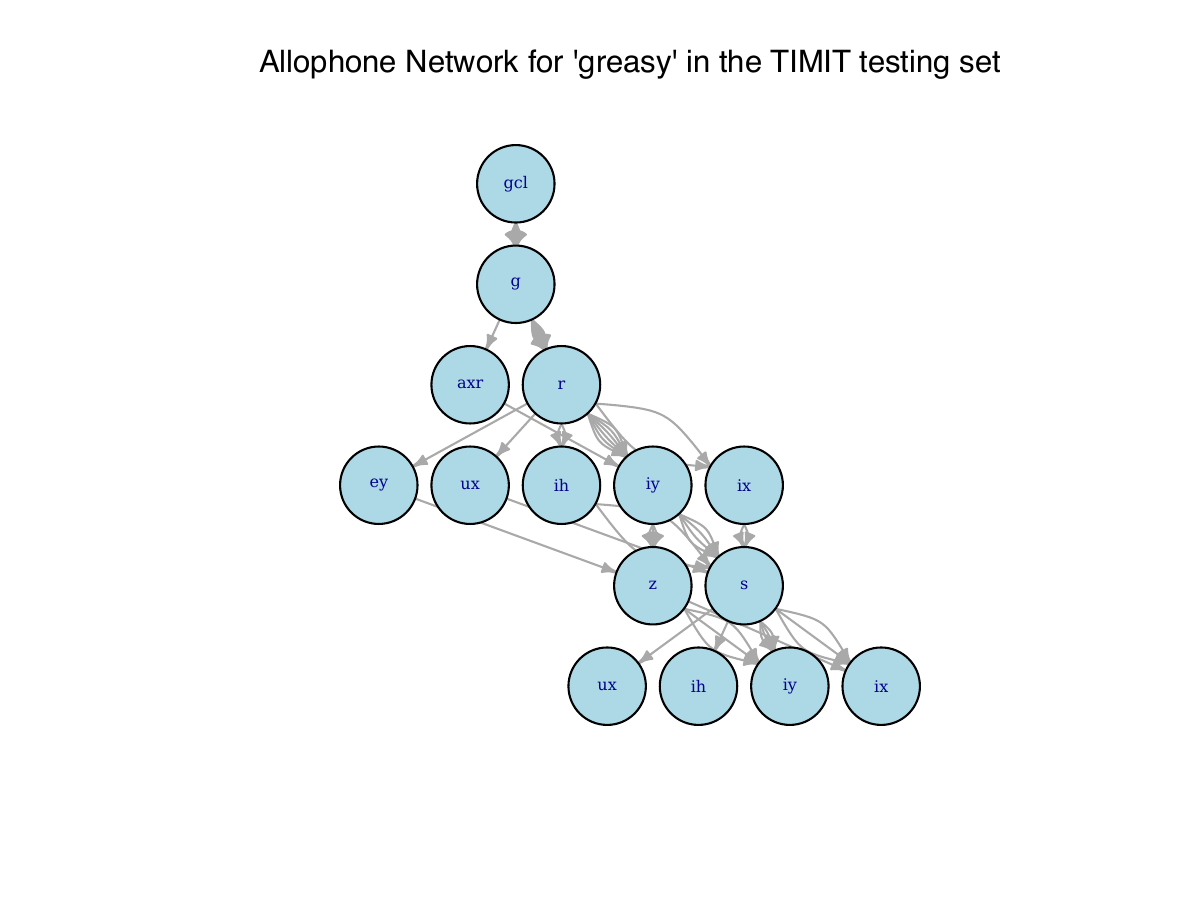}
        \caption{TIMIT phone set representation}
        \label{fig:allophone_network}
    \end{subfigure}
    \hfill
    \begin{subfigure}[t]{0.48\textwidth}
        \centering
        \includegraphics[width=\linewidth]{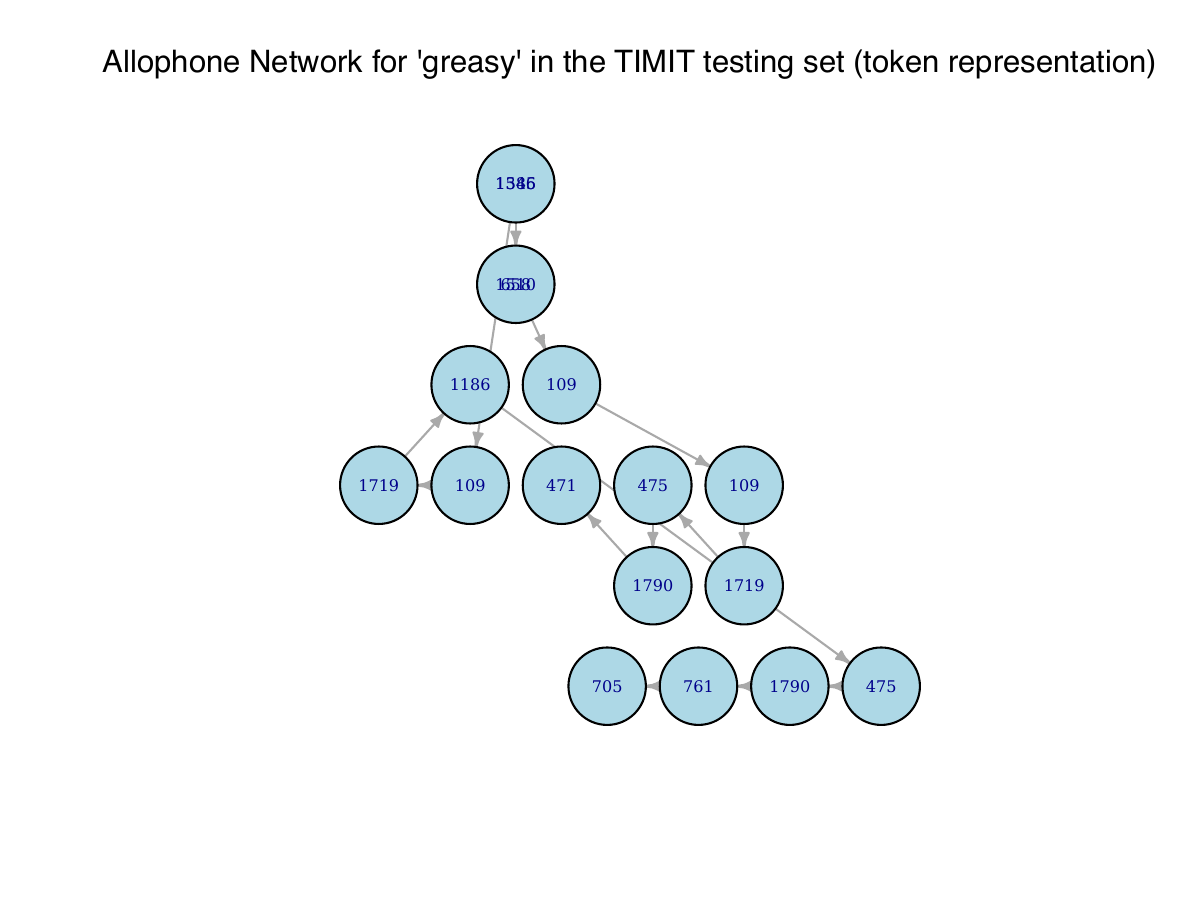}
        \caption{Mimi token representation}
        \label{fig:allophone_network_token}
    \end{subfigure}
    \caption{Comparison of the TIMIT phone transcriptions and Mimi  semantic token representations of the allophone network for the pronunciations of the word \textit{greasy} in the TIMIT training data}
    \label{fig:allophone_networks}
\end{figure*}

The ``allophone network''\cite{cohen1987studies} represented in Figure~\ref{AllophoneNetwork} for the word ``\textit{water}'' shows the visualization of the main allophones expected for this word. A single phone [w] represents the initial consonant. The vowel opposes two possible realizations or trajectories in the network, the ``\textit{bob}'' vowel or the ``\textit{bought}'' vowel. The possibility of flapping is indicated for the medial consonant by DX and two possible allophones are noted for the final vowel, the central rhotic vowel as in \textit{bird} or the reduced vowel with no rhoticity found in northern varieties (schwa). This networking of the allophone variants can be replicated with the word \textit{greasy} using the TIMIT phoneset representations or with the 
neural audio encoder tokens with many more details as evidenced in Figure~\ref{fig:allophone_networks}. Six positions can be observed (compared to 9 in the token/subphonemic representation) for the TIMIT phone set representation of the pronunciations of ``\textit{greasy}'' in the training data. 

\subsection{Speech Tokenizers as Metalinguistic Representations} 
The behaviour of Mimi’s semantic codebook suggests parallels with earlier approaches to subphonetic modeling in speech recognition. Rather than encoding discrete phoneme-like symbols, the tokens appear to capture context-dependent acoustic-phonetic realizations shaped by coarticulation and phonetic environment. This interpretation is consistent with the entropy asymmetries and token-to-phone correspondences observed in our analyses.

A useful comparison can be drawn with senones in Hidden Markov Model (HMM)-based ASR systems \cite{hwang1992subphonetic}. Senones were introduced as context-dependent subphonetic units designed to model variation in phoneme realizations across different acoustic environments. Similarly, Mimi tokens appear to function as distributed acoustic units that encode finer-grained variation than phoneme-level categories alone.

\subsection{Cross-Linguistic Evidence from VoxPopuli}

To extend the analysis beyond English, we examined token distributions across language families using the VoxPopuli corpus \cite{wang2021voxpopuli}. Preliminary observations on Mimi representations across the 27 European Union languages suggest that token distributions tend to cluster according to language families, as illustrated in Table~\ref{tab:language_groups}. This grouping indicates that the first codebook captures structured phonetic variation that is shared across related languages.

\begin{table}[h]
\centering
\begin{tabular}{l c}
\toprule
\textbf{Language Family} & \textbf{Number} \\
\midrule
Germanic    & 85 \\
Romance     & 223 \\
Slavic      & 187 \\
Finno-Ugric & 313 \\
Baltic      & 368 \\
Hellenic    & 520 \\
Semitic     & 378 \\
\bottomrule
\end{tabular}
\caption{Distribution of VoxPopuli languages by family.}
\label{tab:language_groups}
\end{table}

These findings provide additional support for the allophonic hypothesis. If tokens were purely phonemic, one would expect more language-specific and discrete mappings. Instead, the observed clustering across language families suggests that tokens encode phonetic properties that generalize across languages, such as articulatory or acoustic similarities shaped by shared phonological systems.

However, these results also highlight an important limitation: the relationship between phonetic units and Mimi tokens is not strictly one-to-one. Rather than forming a stable inventory of phoneme-like units, the codebook appears to implement a distributed representation in which phonetic entities are mapped onto multiple context-dependent tokens. This observation is consistent with the entropy-based analysis, which shows that each phoneme corresponds to a broad set of competing tokens.

\subsection{Typologically Related Results} 

\begin{figure*}[t]
\centering
\includegraphics[width=0.8\textwidth]{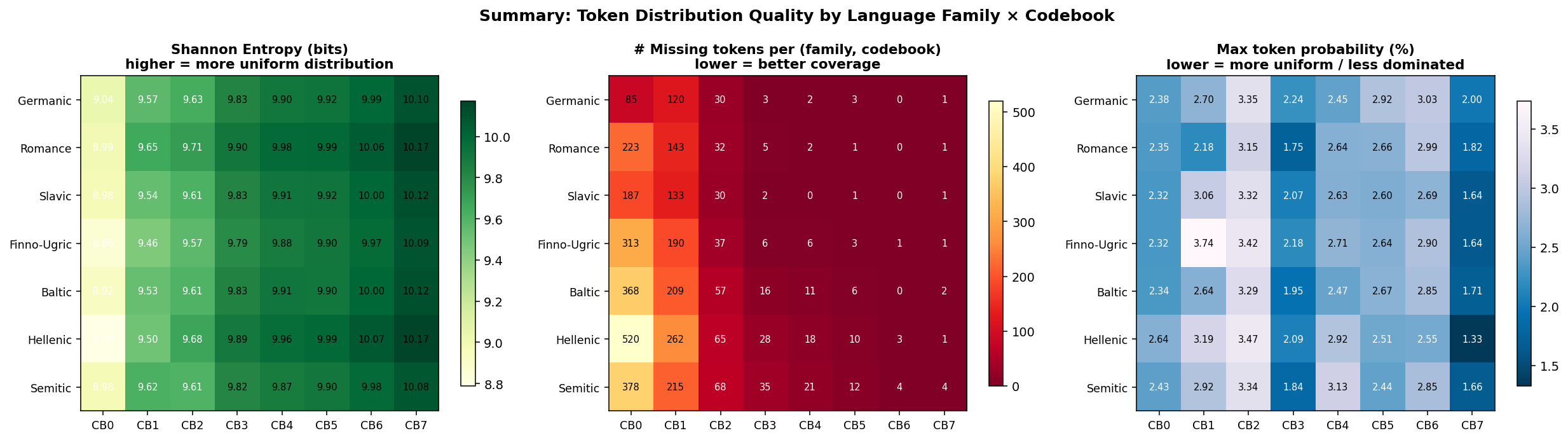}
\caption{Token distribution quality across language families and codebooks. Left: Shannon entropy (higher indicates more uniform distributions). Middle: number of missing tokens (lower indicates better coverage). Right: maximum token probability (lower indicates less dominance).}
\label{fig:voxpopuli_typology}
\end{figure*}

Figure~\ref{fig:voxpopuli_typology} summarizes three complementary metrics: Shannon entropy, token coverage, and maximum token probability, computed for each codebook across typologically related language groups.

First, the entropy patterns are highly consistent across language families. All groups exhibit increasing entropy from codebook\_0 to higher codebooks, indicating progressively more uniform token usage. Crucially, this trend is not language-specific but shared across different families, suggesting that the structure of the representation is largely language-independent.

Second, the number of missing tokens decreases sharply across codebooks, reflecting improved coverage of the acoustic space. While codebook\_0 shows substantial sparsity, higher codebooks achieve near-complete coverage across all language families. This indicates that the representational capacity of the codec is distributed across multiple codebooks rather than localized in a single layer.

Third, the maximum token probability decreases across codebooks, showing that token distributions become less dominated by a small number of units. This again supports the idea of a more distributed and fine-grained encoding at deeper levels of the representation.

Moreover, the fact that token distributions evolve similarly across codebooks for all language families indicates that phonetic structure is not encoded in isolation within codebook\_0, but emerges from the interaction of multiple representational layers. This reinforces the view that Mimi implements a distributed and context-sensitive encoding of speech, in which phonetic units are not mapped one-to-one onto discrete tokens but are instead represented through combinations of tokens across codebooks.

\subsection{Token Distribution Across Codebooks and Language Families}

\begin{figure*}[t]
\centering
\includegraphics[width=0.8\linewidth]{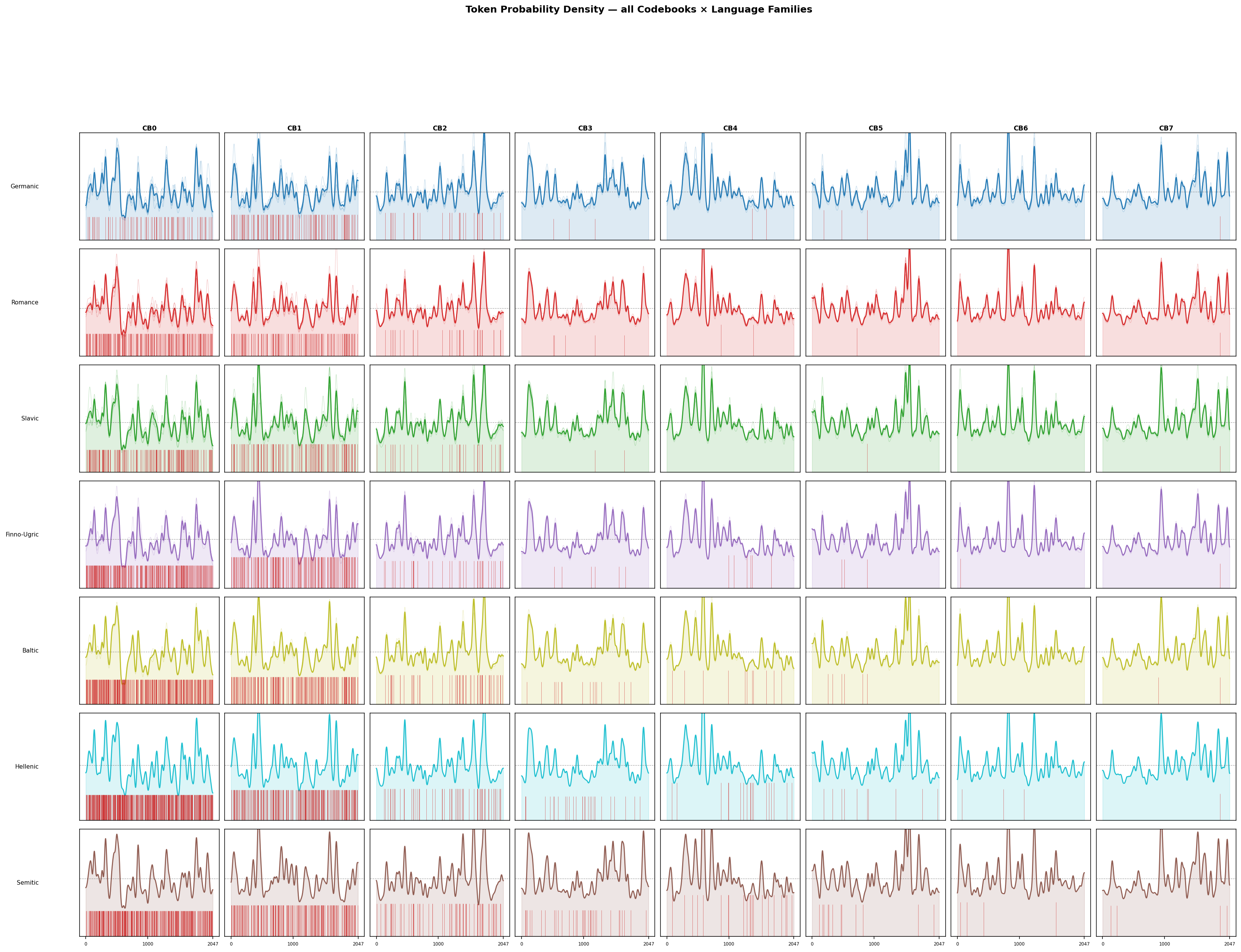}
\caption{Token probability density across codebooks (CB0--CB7) and language families. Each subplot shows the distribution over token indices, highlighting differences in sparsity and dominance across codebooks.}
\label{fig:token_density}
\end{figure*}

To further investigate the structure of the Mimi representations, we analyze the distribution of token probabilities across codebooks and language families. Figure~\ref{fig:token_density} presents the probability density of tokens for each codebook (CB0--CB7) and language family.

A clear structural difference emerges across codebooks. The earliest codebooks (CB0--CB2) exhibit highly peaked distributions, where a small number of tokens dominate the representation. This indicates that these codebooks capture more selective and possibly lower-level acoustic or phonetic features. In contrast, later codebooks (CB4--CB7) display flatter and more uniform distributions, suggesting a broader and less specialized use of tokens.

\subsection{Observations on Greek Phonetic Specificity}

Although phoneme-level alignment is not yet incorporated, preliminary observations can be made regarding Greek, the sole representative of the Hellenic language family in the dataset. Greek is known for its relatively stable vowel system and distinct consonantal contrasts, including fricatives such as /x/ and /\textgamma/, which are less common in many other European languages \cite{mennen2006acquisition}.

In the token distributions, the Hellenic group exhibits patterns broadly consistent with other language families, but with slight variations in peak structure and dispersion. These differences may reflect the influence of language-specific phonetic features, particularly in the realization of fricatives and vowel quality.
\section{Conclusion}
\label{conclusion}
In this paper, we compared the TIMIT phone alignment with Mimi's first codebook representing semantic tokens and measured the consistency of the mapping we produced. It remains to be seen whether the acoustic tokens (the 7 other parallel codebooks used to regenerate speech with the semantic codebook) could  similarly be reconnected to 
properties of the speech signal such as suprasegmental features.
The growing number of downstream tasks relying on this NAC makes it an object of research worthy of interest: the Mimi codec has been used with the Moshi conversational agent and  Hibiki \cite{labiausse2025hibiki} for speech translation. It has more recently been adopted for Sesame's Maya conversational speech model\footnote{\url{https://huggingface.co/sesame/csm-1b}}.

An important limitation of our work is we did not fully investigate gender effect and potential duplicates of tokens, which could be attributed to different formant values for male and female voices. Traditional signal processing methods could be applied to the latent space to extract meaningful features, such as energy, entropy, or frequency content, which might reveal important characteristics of the learned representations. We have not used some of the acoustic correlates of the codes. 
For instance, we could exploit mid-temporal value formant extractions to assign the prototypical value of vowels to their corresponding token IDs.

Finally, these results open a perspective that could be described as computational creolistics: the investigation of how speech representations emerge through the sharing and recombination of tokens across speakers and languages. Rather than forming fixed phonological categories, Mimi tokens appear to function as indexical units, whose distribution reflects both shared acoustic structure and context-dependent variation, paving the way for a computational metalinguistic representation of allophonic classes.

\section*{Acknowledgement}
We thank the three reviewers for their constructive feedback. This publication has emanated from research supported in part by a 2021 research equipment grant from the Scientific Platforms and Equipment Committee (PAPTAN project), under ANR Grant Number ANR-18-IDEX-0001 (Financement IdEx Université de Paris).

\bibliography{custom}

\label{sec:appendix}

\end{document}